\pdfoutput=1

\documentclass[11pt]{article}

\usepackage[final]{acl}

\usepackage{times}
\usepackage{latexsym}

\usepackage[T1]{fontenc}

\usepackage{microtype}

\usepackage{inconsolata}

\usepackage{graphicx}
\usepackage{microtype}
\usepackage{amsmath}
\usepackage{amssymb}
\usepackage{mathtools}
\usepackage{amsthm}
\usepackage{bbm}
\usepackage{subcaption}
\usepackage{multirow}
\usepackage{algpseudocode}
\usepackage{enumitem}
\usepackage{floatrow}
\usepackage{booktabs}

\title{Search-Adaptor: Embedding Customization for Information Retrieval}

\author{Jinsung Yoon, Yanfei Chen, Sercan \"{O}. Ar{\i}k, Tomas Pfister \\
  Google Cloud AI \\
  \texttt{\{jinsungyoon, yanfeichen, soarik, tpfister\}@google.com}}

\begin{document}
\maketitle
\begin{abstract}
Embeddings extracted by pre-trained Large Language Models (LLMs) have significant potential to improve information retrieval and search.
Beyond the zero-shot setup in which they are being conventionally used, being able to take advantage of the information from the relevant query-corpus paired data can further boost the LLM capabilities.
In this paper, we propose a novel method, Search-Adaptor, for customizing LLMs for information retrieval in an efficient and robust way. 
Search-Adaptor modifies the embeddings generated by pre-trained LLMs, and can be integrated with any LLM, including those only available via prediction APIs.
On multiple English, multilingual, and multimodal retrieval datasets, we show consistent and significant performance benefits for Search-Adaptor -- e.g., more than 5\% improvements for Google Embedding APIs in nDCG@10 averaged over 14 BEIR datasets.
\end{abstract}

\section{Introduction}\label{sec:intro}
Information retrieval is broadly considered as the task of searching for information via querying corpus database that might consist many different types of data, such as documents, webpages or logs.
It has a wide range of applications across many industries, including retail, healthcare, and finance, with a significant portion of the world's economy is built on.
Particularly, language modeling is the key part of information retrieval as in most cases, query and corpus data are in text form. 
Large language models (LLMs) have demonstrated significant achievements for a variety of text processing tasks, including question answering, summarization, and mathematical reasoning \citep{brown2020language, chowdhery2022palm, zhang2022opt}. 
One critical enabler for the success on these has been transforming raw text into meaningful representations that preserve semantic meanings in the representation space \citep{ouyang2022training}. 
For a wide range of applications, from recommendations to anomaly detection, tasks are defined as explicit operations on the learned representations.
This makes the quality of the text mapping into embeddings become of paramount importance. 
Information retrieval systems commonly utilize the text embeddings, with relevant corpora being ranked based on the similarity between queries and corpus \citep{wang2022text,izacard2021towards}. 

Various LLMs have been proposed to extract embeddings from raw text, with the notable ones including the Sentence T5 \citep{ni2021sentence}, OpenAI Embedding APIs \cite{openai-text-embedding} and Google Embedding APIs \cite{gcp-text-embedding}.
However, one fundamental limitation of pre-trained LLMs is that they cannot utilize retrieval-specific target task data, that are in the form of positively relevant query-corpus pairs.
Even with a small amount, using such data for tuning is expected to significantly improve information retrieval capabilities. 
Conventional fine-tuning \citep{howard2018universal} can be one straightforward way of utilizing the paired query-corpus information. 
However, if the number of paired samples is small, tuning all the weights of a model might yield overfitting and thus poor generalization \citep{lin2023speciality}, especially in the presence of distribution shifts.
In addition, it can be costly from a computational perspective as it requires large memory.
There are multiple parameter-efficient tuning methods such as prompt tuning \citep{lester2021power,li2021prefix}, LoRA \citep{hu2021lora}, partial fine-tuning \citep{zaken2021bitfit}, and various adapters \citep{houlsby2019parameter,ruckle2020adapterdrop}. 
These approaches only tune a subset of the parameters of LLMs, aiming to reduce the risks of overfitting and bringing computational gains. 
As a common bottleneck, all of these methods need full access to the LLM's parameters to tune the model, which may not be possible for LLMs with API-based inference-only access.

\begin{figure}[t!]
    \centering
    \includegraphics[width=\textwidth]{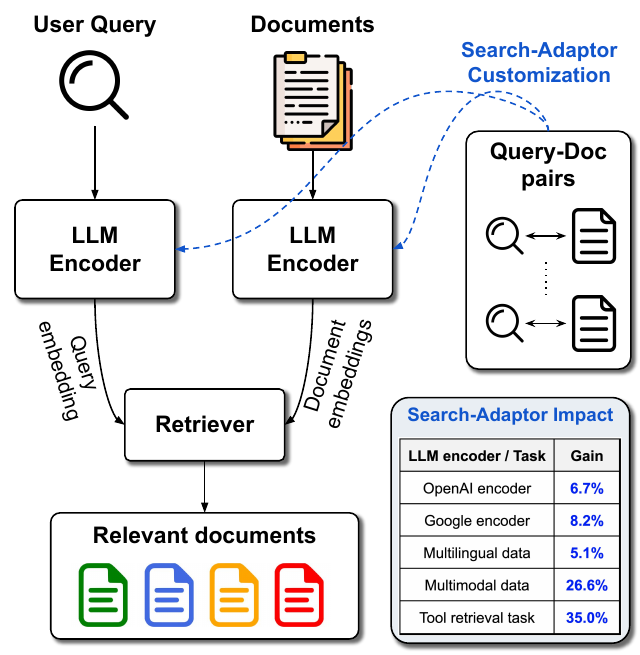}
    \caption{Search-Adaptor modifies the pre-trained LLM embeddings by learning from paired query-document data, yielding significantly improved retrieval performance on target tasks.
    Note that Search-Adaptor does not require access to the LLM weights or gradients and can even be applied to the LLMs that are only accessible via prediction APIs.}
    \label{fig:high-level-figure}
\end{figure}

In this paper, our focus is customizing LLMs to obtain superior embeddings for information retrieval, particularly from the angle of how to best take advantage of the available retrieval-specific tuning data and obtain robust improvements in wide range of regimes, with a tuning method that is low cost and even applicable to LLMs with API-based inference-only access. 
Fig.~\ref{fig:high-level-figure} overviews the proposed approach, Search-Adaptor.
Improving the tuning performance in this setup requires a set of innovations. 
We propose integration of a low capacity adapter module (to be customized for the target dataset) on top of fixed LLMs to modify the pre-trained embeddings.
For efficient tuning, we introduce a novel differentiable ranking loss that can directly utilize the information of positive query and corpus pairs. 
In addition, we include multiple regularizers to improve generalization in this small data regime where without intervention, the pre-trained LLMs would end up with catastrophic forgetting.
Enabled by such design approach, one major advantage of Search-Adaptor is that it does not require access to the parameters of the pre-trained LLMs -- only the inference outputs of the LLMs are needed. 
Commercial embedding APIs that show state-of-the-art performance usually do not provide access to their model parameters. 
In such cases, Search-Adaptor can still be used to further improve those API-based embedding models, in contrast to alternative tuning methods.
We demonstrate the effectiveness of Search-Adaptor across 14 BEIR datasets \citep{thakur2021beir} and 17 MIRACL multilingual datasets \citep{zhang2022making} with Google and OpenAI embedding APIs, applying the Search-Adaptor on top. 
In addition, we evaluate Search-Adaptor's performance improvements with open-source Sentence T5 models \citep{ni2021sentence}.
Overall, Search-Adaptor provides significant improvements over alternatives, consistently across different datasets and models. 
The contributions of this paper are:
\begin{itemize}
    \item We propose a novel adaptation framework for information retrieval applications that can significantly improve the pre-trained LLMs.
    \item We introduce a novel ranking loss and multiple regularizers that reduce overfitting and forgetting and thereby improve the retrieval performance even in the small data regime.
    \item We provide consistent and significant improvements for retrieval performance with a range of datasets (from 1K to 500K positive query-corpus training data pairs).
    \item We show that Search-Adaptor on smaller LLMs can approach the performance of larger LLMs in zero-shot setting, underlining its potential for model distillation. 
    \item We extend the application of Search-Adaptor to multimodal learning and tool use scenarios, demonstrating its significant benefits.
\end{itemize}

\section{Related Work}\label{sec:related_works}
\textbf{Pre-trained LLMs for zero-shot retrieval.} 
LLMs to extract general text embeddings are commonly used in both academia and industry. 
AI solution providers like Google \cite{gcp-text-embedding} and OpenAI have productionized general text embeddings that can be directly used via simple APIs for zero-shot retrieval applications. 
In addition, many previous efforts have introduced new general text embedding models with various pre-training methods and datasets. 
GTE \cite{li2023towards} proposes a multi-stage pre-training of embedding models with diverse naturally paired text datasets. 
E5 \cite{wang2022text} pre-trains the embedding models by weakly-supervised contrastive learning, utilizing consistency-based filter to generate high quality text pairs for pre-training. 
Note that Search-Adaptor can be applicable on top of any pre-trained LLM embedding models to customize their embeddings for superior retrieval performances. 

\textbf{Embedding customization.} 
Instead of using one unified model for zero-shot retrieval, the embeddings can be customized for each dataset or task.
Instruction-based embedding customization is one popular method. 
TART \cite{asai2022task} builds a retrieval system that adapts the retrieval based on the instruction. 
Different retrieval tasks (e.g., code, question, or answer) are given as the instruction to further improve dense embedding retrieval. 
InstructOR \cite{su2022one} integrates the task and domain descriptions prior to the input to fine-tune the embeddings for retrieval. 
However, these do not directly utilize the provided relevant query-corpus pairs. 
Full or parameter-efficient fine-tuning (such as LoRA \cite{hu2021lora} and (IA)$^3$ \cite{liu2022few}) can also be considered for embedding customization. 
Pre-trained LLMs can be fine-tuned with contrastive loss using positive query-corpus paired data. 
Promptagator \cite{dai2022promptagator} utilizes in-context learning to generate synthetic query-corpus pairs using a few number of original query-corpus pairs, and subsequently using those synthetic pairs to fine-tune the pre-trained LLMs.
However, all these are only applicable when there is full access to the parameters of pre-trained LLMs, which is often not possible for state-of-the-art commercial text embedding models. 
On the other hand, Search-Adaptor can be applied without full access to the LLM parameters.

\section{Problem Formulation}\label{sec:problem}
We formulate the retrieval problem with a given query-corpus paired dataset. 
Assume a query set denoted as $\mathcal{Q} = \{q_1, ..., q_N\} \in Q$ and a corpus set denoted as $\mathcal{C} = \{c_1, ..., c_M\} \in C$.
Each positive relationship between a query and corpus is represented as the triplet $r_{ij} = (q_i, c_j, y_{ij})$ with $y_{ij} > 0$ as the strength of the relationship between $q_i$ and $c_j$. 
We treat all other triplets as negative relationships ($y_{ij} = 0$). 
The set of all query-corpus relationships is denoted as $\mathcal{R} = \{(q_i, c_j, y_{ij})\}_{i=1:N, j=1:M} = \mathcal{R}_p \cup \mathcal{R}_n$, where $\mathcal{R}_p = \{(q_i, c_j, y_{ij}) \in \mathcal{R} | y_{ij} > 0\}$ is the set of positive relationships and $\mathcal{R}_n = \{(q_i, c_j, y_{ij}) \in \mathcal{R} | y_{ij} = 0\}$ is the set of negative relationships. 
Note that $y_{ij}$ can be either binary or continuous.

The retrieval system aims to find the relationship between the given query ($q_i$) and corpus ($c_j$) such that the predicted relationship is highly correlated with the ground truth relationship ($y_{ij}$).
The scoring function $f: Q \times C \rightarrow \mathbb{R}$ takes queries and corpus data as inputs and outputs a score estimate on the relationship between them. 
The optimal score is the one that has the same order as the ground truth relationship for each query.

\begin{figure}[t!]
    \centering
    \includegraphics[width=\textwidth]{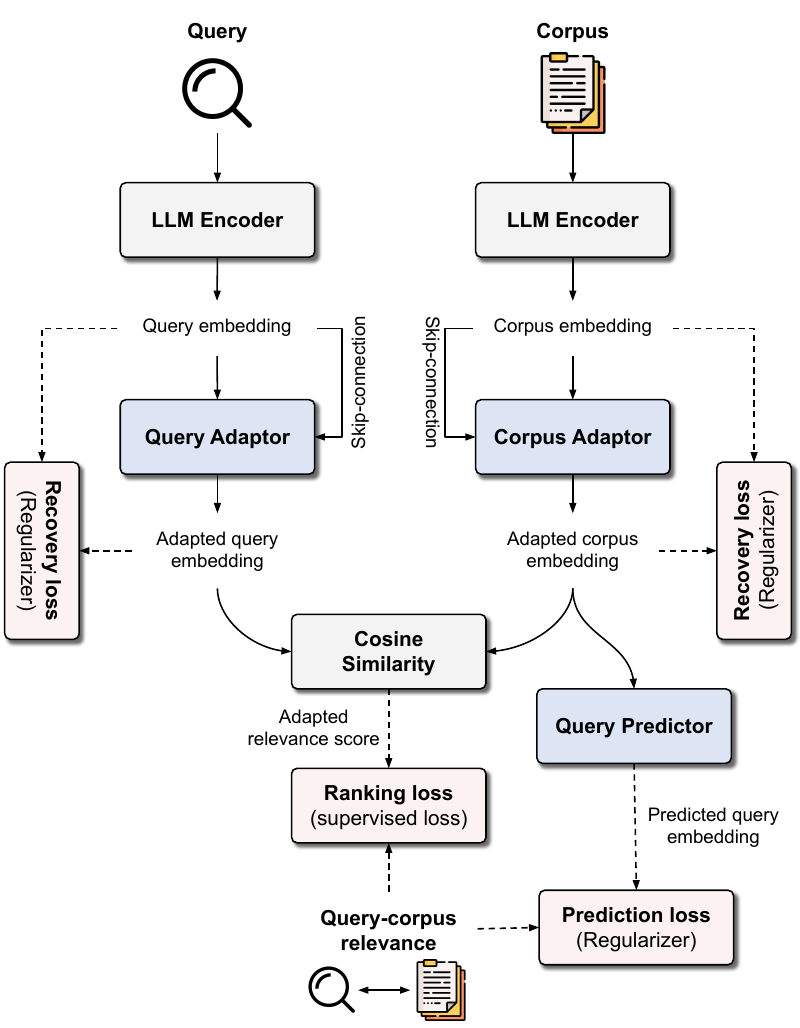}
    \caption{Block diagram of Search-Adaptor. Grey colored blocks are fixed components (e.g., a text embedding API); blue-colored blocks are additional trainable building blocks; and red-colored blocks are for loss computations. At inference, only query and corpus adapters are utilized and the query predictor can be discarded.}
    \label{fig:search-adaptor-block-diagram}
\end{figure}

\section{Methods}\label{sec:method}
Fig.~\ref{fig:search-adaptor-block-diagram} overviews the components of  Search-Adaptor, that are described in following sections. 

\subsection{Adapting pre-trained LLMs}
Major real-world constraints for tuning the LLM embedding models shape our methodological design.
The tuning operation for high-capacity models can be very costly, and one often does not have access to the parameters and the gradients of the pre-trained models (e.g. LLMs with API-based inference-only access).
This motivates the need for an adaptation method that can operate with fixed pre-trained embedding models with an efficient adaptation module, to extend to the LLMs with API-based inference-only access.
In Search-Adaptor, we propose modifying the embeddings extracted from pre-trained LLMs for superior search and information retrieval.

Consider the query and corpus embeddings extracted using the pre-trained embedding model $E$: $\mathcal{Q}_E = \{qe_1, ..., qe_N\} \in \mathbb{R}^d$ and $\mathcal{C}_E = \{ce_1, ..., ce_N\} \in \mathbb{R}^d$ where $qe_i = E(q_i)$ and $ce_j = E(c_j)$. 
Note that both query and corpus embeddings are in the same embedding space.

The objective of Search-Adaptor is to modify embeddings extracted from pre-trained LLMs in a way that maximizes retrieval performance. 
A learnable adaptation function is defined as $f: \mathbb{R}^d \rightarrow \mathbb{R}^d$, which maps the original embedding to a new embedding that is more favorable for retrieval applications. 
The modified embeddings are denoted as $\hat{\mathcal{Q}}_E = \{\hat{qe}_1, ..., \hat{qe}_N\} \in \mathbb{R}^d$ and $\hat{\mathcal{C}}_E = \{\hat{ce}_1, ..., \hat{ce}_M\} \in \mathbb{R}^d$ where $\hat{qe}_i = f(qe_i)$ and $\hat{ce}_j = f(ce_j)$. 
The relevance scores between modified query and corpus embeddings are defined as follows:
\begin{equation*}
    \hat{s}_{ij} = \text{Cosine-Similarity}(\hat{qe}_i, \hat{ce}_j) = \frac{\hat{qe}_i \cdot \hat{ce}_j}{||\hat{qe}_i|| ||\hat{ce}_j||}.
\end{equation*}
Search-Adaptor consists of the following components (see Fig.~\ref{fig:search-adaptor-block-diagram} for details):
\begin{itemize}
    \item \textbf{Adaptation function $f$}. This function is used to modify the query and corpus embeddings. We add a skip connection to $f$ so that it can only learn the residual between the original and adapted embeddings as follows: $\hat{qe}_i = qe_i + f(qe_i)$ and $\hat{ce}_j = ce_i + f(ce_i)$. Note that we use the shared adapter for both query and corpus (see Sec.~\ref{sec:discussion} for ablation studies). The ranking loss, reconstruction loss, and prediction loss are used to train $f$.
    \item \textbf{Query predictor $p$}. This function is used to predict the query embedding using the adapted corpus embedding. The prediction loss is used to train $p$.
\end{itemize}
At inference, we only use the adaptation functions ($f$) to modify the query and corpus embeddings. 
We then compute the cosine similarity between the modified query and corpus embeddings to estimate the relevance between query and corpus. 
Query predictor is not used at inference. 

\subsection{Ranking objective}
As explained in Sec.~\ref{sec:problem}, the objective of the retrieval is to predict the correct order of the relevance between queries and corpus. 
Therefore, the most critical part is to properly design the ranking loss. 
We propose a ranking loss as follows:
\begin{align*}
    \mathcal{L}_{Rank} = \sum_{i=1}^N \sum_{j=1}^M \sum_{k=1}^M  & I(y_{ij} > y_{ik}) \cdot (y_{ij} - y_{ik}) \\
    &\cdot \log(1+e^{(s_{ik} - s_{ij})}),
\end{align*}
where $I(y_{ij} > y_{ik})$ is an indicator function that is equal to $1$ if $y_{ij} > y_{ik}$ and $0$ otherwise.
$s_{ij}=\text{Cosine-Similarity}(E(q_i), E(c_j))$ is the relevance score between query text ($q_i$) and corpus text ($c_j$). 

The ranking loss penalizes the model more when it predicts a lower score for a pair of query and corpus that has a higher ground truth relevance (i.e., $s_{ij} < s_{ik}$ even though $y_{ij} > y_{ik}$). 
The amount of penalization is proportional to the difference in ground truth relevance $(y_{ij} - y_{ik})$ and the difference in estimated scores $\log(1+e^{(s_{ik} - s_{ij})})$.
Note that $\log(1+e^{(s_{ik} - s_{ij})})$ can be replaced with any monotonic function such as linear function.
In general, the ranking loss encourages the model to predict higher scores for pairs of query and corpus that have a higher ground truth relevance.
Table~\ref{tab:ablation_study} shows a comparison of this ranking loss to alternatives and demonstrates its effectiveness.

\subsection{Regularization}
Introducing proper inductive biases via regularization is important to improve adaptation from pre-trained LLM embeddings without forgetting too much information from the pre-trained LLMs. 
Towards this end, we propose two regularization methods:

\textbf{Recovery.} To increase generalizability, we postulate avoiding modification of the adapted embedding too far away from the original embedding. 
As such, we propose minimization of the difference between the original and adapted embeddings using a recovery regularizer:
\begin{equation*}
    \mathcal{L}_{Rec} = \frac{1}{N} \sum_{i=1}^N||\hat{qe}_i - qe_i||_1 + \frac{1}{M} \sum_{j=1}^M||\hat{ce}_i - ce_i||_1
\end{equation*}
where $\hat{qe}_i$ is the adapted query embedding and ${qe}_i$ is the original query embedding. Similarly, $\hat{ce}_i$ is the adapted corpus embedding and $ce_i$ is the original corpus embedding.
The recovery regularizer encourages the adapted embeddings to be not too far from the original embeddings. 

\textbf{Prediction.} 
Intuitively, if the query and corpus are highly relevant, we can use the corpus to predict the query. 
Building upon this intuition, we propose a regularizer in the form of prediction loss between the query and corpus, calculated as follows:
\begin{equation*}
    \mathcal{L}_{Pred} = \frac{\sum_{i=1}^N \sum_{j=1}^M y_{ij} \cdot ||\hat{qe}_i - p(\hat{ce}_j)||_1}{\sum_{i=1}^N \sum_{j=1}^M y_{ij}}
\end{equation*}
where $p: \mathbb{R}^d \rightarrow \mathbb{R}^d$ is a function that predicts the query given the corpus, and $y_{ij}$ is a weight that is assigned to the loss if the query and corpus are correlated. 
The prediction loss encourages the model to predict the query well given the corpus, especially if the query and corpus are correlated. 
Note that we do not include the prediction function from query to corpus because usually corpora are significantly longer than queries which would render the task challenging.

\subsection{Training}
Using the proposed ranking loss, recovery loss, and prediction loss, we optimize the adaptation function $f$ and prediction function $p$ by minimizing the following loss function:
\begin{equation*}
    \min_{f,p} \mathcal{L}_{Rank}(f) + \alpha \mathcal{L}_{Rec}(f) + \beta \mathcal{L}_{Pred}(f, p),
\end{equation*}
where $\alpha \geq 0$ and $\beta \geq 0$ are the hyper-parameters that control the relative importance of the different loss terms.\footnote{In the experiments, we tune these hyper-parameters based on validation set ($\alpha \in \{0.0,0.1,1.0\}$ and $\beta \in \{0.0,0.01,0.1\}$).} 
Table~\ref{tab:ablation_study} shows the results of ablation studies on the effectiveness of the different loss terms. 
All hyper-parameters are provided in Appendix~\ref{appx:hyperparams}.

Note that the ranking loss compares all possible pairs between queries and corpus which needs $NM^2$ times computations per one epoch ($M >> N$). 
For efficient computation, we randomly subsample the corpus for each query batch. 
While doing so, we always include the corpus which has positive relevance to queries in that batch.



\section{Experiments}
We evaluate the performance of Search-Adaptor in multiple scenarios on numerous datasets. 
We demonstrate that Search-Adaptor is model-agnostic, applying it both on top of API-based LLMs (merely via access to Google \& OpenAI APIs) and open-sourced LLMs (e.g., Sentence T5 \citep{ni2021sentence}). 
We also demonstrate that it is data-agnostic by evaluating Search-Adaptor on English, multilingual and multimodal datasets.

\subsection{Experimental settings}
We first consider the 14 retrieval datasets from the BEIR repository \citep{beir_dataset} to evaluate the performance in English data, with corpus sizes ranging from 3.6K to 8.8M, and training pairs ranging from 0.7K to 532K. 
For the datasets with only test data (e.g., Arguana, SciDocs), we split the data into disjoint train and test sets with a 50/50 ratio, based on the sorted query IDs. 
We also use MIRACL data \citep{miracl_dataset} which consists of 17 multilingual datasets including Japanese, Chinese, French, and Indonesian. 
More dataset details can be found in Appendix~\ref{appx:data}.

We use nDCG@10 as the main metric to quantify the retrieval performance (see Appendix~\ref{appx:metric} for more details). 
For model selection, we divide the training data into disjoint training and validation splits with an 80/20 ratio, and select the model with the highest validation nDCG@10 value.

We consider both API-based and open-sourced LLMs. 
As the API-based LLM, we use OpenAI embedding API \cite{openai-text-embedding} and Google embedding API \cite{gcp-text-embedding}. 
As the open-sourced LLM, we use Sentence T5 models\footnote{\url{https://tfhub.dev/google/sentence-t5/st5-base/1}} of two different sizes, GTE-Large \cite{li2023towards}, GTR-Large \cite{ni2021large} and Condenser-Retriever \cite{gao2021condenser}.

\subsection{Adapting with API-based LLMs}\label{sec:experiment1}

\begin{figure*}[t!]
    \centering
    \includegraphics[width=\textwidth]{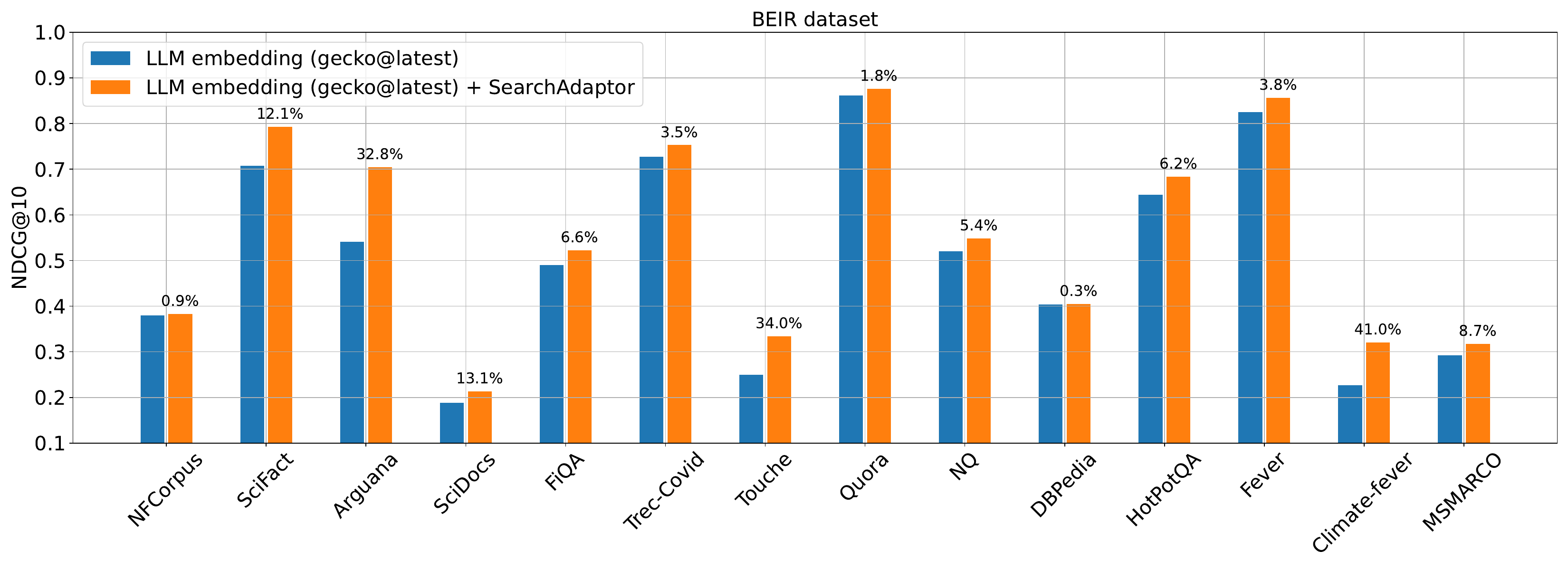}
    \caption{Performance improvements with Search-Adaptor on top of Google's LLM based embedding APIs (gecko@latest, 768 dimensions) for 14 BEIR datasets.}
    \label{fig:API-based-result-beir}
\end{figure*}

\begin{figure*}[t!]
    \centering
    \includegraphics[width=\textwidth]{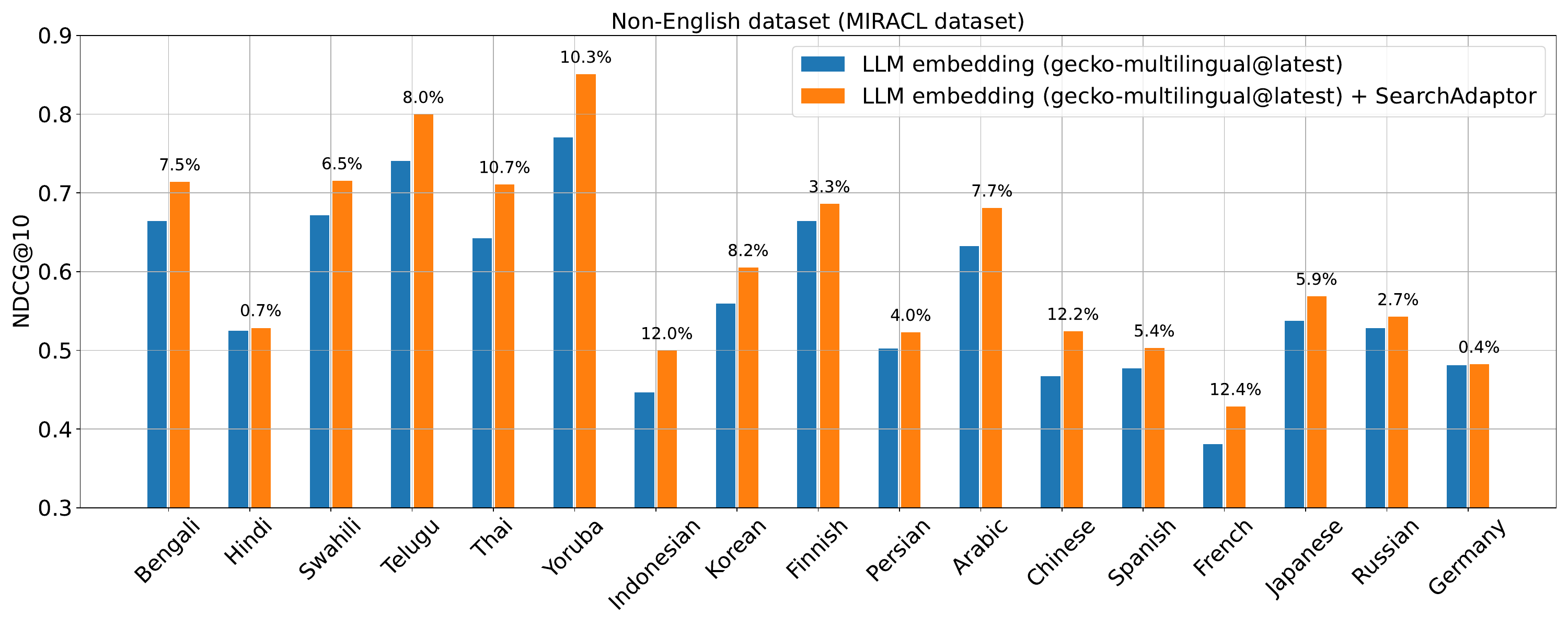}
    \caption{Performance improvements with Search-Adaptor on top of the Google's LLM based embedding APIs (gecko-multilingual@latest, 768 dimensions) for 17 multilingual MIRACL datasets.}
    \label{fig:API-based-result-non-english}
\end{figure*}

One of the biggest advantages of Search-Adaptor is that it can be applied on top of any API-based LLM -- without having access to the parameters of LLMs, Search-Adaptor can further improve the retrieval performance. 
This is particularly important as the state-of-the-art LLMs are actually API-based models (owned by companies). 

\begin{table}[h!]
    \centering\small
    \begin{tabular}{c|c|c|c}
    \toprule
        \multirow{2}{*}{Datasets}  & {Zero-} & Search- & OpenAI's \\
        & shot & Adaptor & Solution \\
        \midrule
        NFCorpus  & 0.3750  & \textbf{0.3785} & 0.2595 \\
        SciFact  & 0.7221  & \textbf{0.7904} & 0.6449 \\
        Arguana & 0.5885 & \textbf{0.6493} & 0.6151 \\
        SciDocs & 0.2003 & \textbf{0.2158} & 0.1941 \\
        FiQA  &{0.4366} & \textbf{0.4410} &  0.3119 \\
        Trec-Covid & 0.7224 & \textbf{0.7733}& 0.7712 \\
        Touche & 0.2590 & \textbf{0.3312} & 0.3157 \\
        Quora  & {0.8830} & \textbf{0.8869} & 0.8670 \\
        \bottomrule
    \end{tabular}
    \caption{Performance improvements with Search-Adaptor and OpenAI's embedding customization solution (OpenAI's solution) on top of OpenAI's LLM based embedding APIs (text-embedding-ada-002, 1536 dimensions) with 8 BEIR datasets.}
    \label{tab:API-based-result-beir-openai}
\end{table}

As can be seen in Fig.~\ref{fig:API-based-result-beir} and Table~\ref{tab:API-based-result-beir-openai}, we demonstrate the retrieval performance improvements on top of API-based text embedding models across 14 datasets from the BEIR repository. 
On average, Search-Adaptor achieves 0.0475 and 0.0349 nDCG@10 improvements for both Google and OpenAI text embedding APIs.
The improvements of some datasets are quite significant indeed -- e.g., 0.1739 with Arguana, 0.0856 with Scifact datasets.

We also compare with OpenAI’s embedding customization solutions (OpenAI's solution)\footnote{\url{https://github.com/openai/openai-cookbook/blob/main/examples/Customizing_embeddings.ipynb}}. Table~\ref{tab:API-based-result-beir-openai} shows worse performance compared to not only Search-Adaptor but also Zero-shot OpenAI embedding in terms of ranking metrics (nDCG@10). 
OpenAI's solution tries to solve classification problems between positive and negatively correlated text pairs. 
It adds “random negatives” to their positive paired samples and try to distinguish between positive and “random negative” pairs using MSE loss. 
This problem is much easier than the retrieval problem (that Search-Adaptor tries to solve), where the task is to identify the positive pairs from all possible negative pairs (including hard negatives). 
Also, Search-Adaptor utilizes ranking loss while OpenAI's solution utilizes MSE loss (which is beneficial for classification or regression problems). 
Lastly, OpenAI's solution does not have regularizations that mitigate the high risks of overfitting when the number of query-corpus pairs is small.

\begin{table*}[t!]
    \centering
    \small
    \begin{tabular}{c|c|c|c||c|c|c}
    \toprule
        Base models & \multicolumn{3}{c||}{ST5-Base} & \multicolumn{3}{c}{GTE-Large} \\
        \midrule
        Datasets & Zero-shot & Search-Adaptor & Full Fine-tuning  & Zero-shot & Search-Adaptor & LoRA \\
        \midrule
        NFCorpus & 0.3100 & 0.3258  & \textbf{0.3501} & 0.3810 & \textbf{0.4063} & 0.3512 \\
        SciFact & 0.5237 & 0.7255  & \textbf{0.7542} & 0.7419 & \textbf{0.8179} & 0.6433 \\
        Arguana & 0.3646 & 0.5501  & \textbf{0.6239} & 0.5987 & \textbf{0.6292} & 0.6091 \\
        SciDocs & 0.1393 & \textbf{0.1657}  & 0.1640 & 0.2460 & \textbf{0.2531} & 0.2209 \\
        FiQA & 0.4064 & 0.4416  & \textbf{0.4557} & 0.4362 & \textbf{0.4428} & 0.4328 \\
        Trec-Covid & 0.5990 & \textbf{0.6986}  & 0.4178 & 0.7242 & 0.7593 & \textbf{0.7656} \\
        Touche & 0.2291 & \textbf{0.3393} & 0.1844 & 0.2566  & \textbf{0.2905} & 0.2752 \\
        Quora & 0.7484 & \textbf{0.8664} & 0.7817 & 0.8831 & 0.8842 & \textbf{0.8871} \\
        \midrule
        Average & 0.4151 & \textbf{0.5141} & 0.4151 & 0.5335 & \textbf{0.5604} & 0.5232\\
        \bottomrule
    \end{tabular}
    \caption{Performance comparison with Search-Adaptor and fine-tuning baselines on top of open-sourced embedding models (ST5-Base and GTE-Large) with 8 BEIR datasets.}
    \label{tab:Open-sourced-result}
\end{table*}


\subsubsection{Search-Adaptor on multilingual data}
Search-Adaptor is also applicable on non-English multilingual data. 
In Fig.~\ref{fig:API-based-result-non-english}, Search-Adaptor shows consistent performance improvements on top of Google Embedding API across 17 different languages (on average 0.0396 nDCG@10 improvement). 
For some languages, it is particularly significant, e.g. the improvement is 0.0687 for Thai. 
These highlight Search-Adaptor being a model-agnostic and data-agnostic approach. 
More experiments with additional embedding models (e.g., GTR-Large \cite{ni2021large} and Condenser-Retriever \cite{gao2021condenser}) can be found in Appendix~\ref{appendix:additional_base_embedding_model}. 
Qualitative analyses can be also found in Appendix~\ref{appendix:qualitative_analysis_sa}.

\subsection{Adapting with open-sourced LLMs}\label{sec:experiment1}
Beyond API-based LLMs, Search-Adaptor can be applied to open-sourced LLMs. 
Here, we consider Sentence T5-Base \cite{ni2021sentence} and GTE-Large \cite{li2023towards} models as the open-sourced LLMs to demonstrate the performance improvements over the baselines.

As shown in Table~\ref{tab:Open-sourced-result}, Search-Adaptor shows consistent improvements over zero-shot ST5-Base model. 
For the open-sourced LLMs, we can also utilize full fine-tuning (with contrastive loss) and LoRA \cite{hu2021lora} as alternatives of Search-Adaptor, albeit the higher training cost. 
The experimental results on the considered benchmarks show that on average, full fine-tuning and LoRA performances can indeed be worse than Search-Adaptor. 
Surprisingly, the performance of full fine-tuning and LoRA can even be much worse than the zero-shot baseline (e.g., for Trec-Covid, Touche with full fine-tuning and NFCorpus, SciFact with LoRA) which is attributed to fine-tuning being prone to overfitting and poor generalization \citep{lin2023speciality}.

With a limited number of query-corpus pairs, Search-Adaptor performs better than fine-tuning methods due to lower risk of overfitting.
Also, training cost (both memory and computations) is much lower with Search-Adaptor than fine-tuning methods.
On the other hand, with enough query-corpus pairs, fine-tuning methods may perform better than Search-Adaptor with open-source retrievers.
Based on these pros and cons, the appropriate customization method can be selected given the specific application scenario. 
If black-box retrievers work better than open-source retrievers or when the number of query-corpus pairs is small, then Search-Adaptor would be a superior choice of customization. 
On the other hand, if open-source retrievers work better than black-box retrievers and the number of query-corpus pairs is large, we can utilize fine-tuning methods to customize their retrievers.

\subsection{Search-Adaptor with multimodal data}

\begin{table}[h!]
    \centering\small
    \begin{tabular}{c|c|c|c}
    \toprule
        \multirow{2}{*}{Datasets}  & {Zero-} & Search- & Gains \\
        & shot & Adaptor &(\%) \\
        \midrule
        Dresses  & 0.2315  & \textbf{0.2681} & 15.8\% \\
        Jackets  & 0.1652  & \textbf{0.2319} & 40.4\% \\
        Pants & 0.1248 & \textbf{0.1821} & 45.9\%\\
        Skirts & 0.1923 & \textbf{0.2282} & 18.7\%\\
        Tops  & 0.2270 & \textbf{0.2542} & 12.0\% \\
        \bottomrule
    \end{tabular}
    \caption{Multimodal retrieval performance (text to image) with Search-Adaptor for Google Cloud's LLM based multimodal embedding API (1408 dimensions) with Fashion-200K dataset.}
    \label{tab:multimodal_result}
\end{table}

\begin{table*}[h!]
    \centering
    \small
    \begin{tabular}{l|c|c|c|c|c|c}
    \toprule
        Variants & NFCorpus & SciFact & Arguana & SciDocs & FiQA & Trec-covid  \\
        \midrule
        Zero-shot baseline & 0.3100 & 0.5237 & 0.3646 & 0.1393 & 0.4064 & 0.5990 \\
        Original Search-Adaptor & \textbf{0.3258} & \textbf{0.7255} & \textbf{0.5501} & \textbf{0.1657} & \textbf{0.4416} & \textbf{0.6986}  \\
        \midrule
        \multicolumn{7}{c}{\textbf{(a) Architectural modifications}}\\
        \midrule
        Without skip connection  & 0.3243 & 0.6465 & 0.5110 & 0.1579 & 0.4133 & 0.6380 \\
        With separate adapters  & 0.3047 & 0.5488 & 0.3659 & 0.1463 & 0.3977 & 0.6148 \\
        \midrule
        \multicolumn{7}{c}{\textbf{(b) Regularizer modifications}}\\
        \midrule
        Without prediction loss  & 0.3235 & 0.6501 & 0.5456 & 0.1642 & 0.4078 & 0.6177 \\
        Without reconstruction loss  & 0.3245 & 0.6491 & 0.5439 & 0.1637 & 0.4127 & 0.6551 \\
        \midrule 
        \multicolumn{7}{c}{\textbf{(c) Alternative ranking losses}}\\
        \midrule
        Sigmoid cross entropy  & 0.3026 & 0.5917 & 0.4912 & 0.1567 & 0.4052 & 0.6702 \\
        Contrastive loss \citep{izacard2021unsupervised} & 0.3046 & 0.5316 & 0.4822 & 0.1449 & 0.4091 & 0.6723  \\
        Softmax cross entropy \citep{bruch2019analysis} & 0.3097 & 0.5452 & 0.4874 & 0.1346 & 0.4121 & 0.6549 \\
        RankNet loss \citep{burges2005learning}  & 0.3119 & 0.5511 & 0.4699 & 0.1599 & 0.4155 & 0.6428 \\
        \bottomrule
    \end{tabular}
    \caption{Ablation studies with variants of Search-Adaptor. As ablation scenarios, we modify regularizers and architectures of the original Search-Adaptor, and replace the proposed ranking loss with alternative ranking losses.}
    \label{tab:ablation_study}
\end{table*}

Search-Adaptor makes consistent and significant improvements when applied on text embeddings. 
We also extend Search-Adaptor from text-only to multimodal data, with image and text, using Google Cloud's multimodal embedding API \footnote{\url{https://cloud.google.com/vertex-ai/generative-ai/docs/embeddings/get-multimodal-embeddings}}. 
We use the Fashion-200K dataset \cite{han2017automatic} for the text to image retrieval task to show how much Search-Adaptor can make further improvements on top of base multimodal embedding API.

Table~\ref{tab:multimodal_result} shows that Search-Adaptor can achieve 20-30\% of performance improvement in nDCG@10 across 5 sub datasets of Fashion-200K datasets. Relevant qualitative analyses can be found in Appendix~\ref{sect:qualitative_multimodal}.

\subsection{Search-Adaptor for tool retrieval}

\begin{table}[h!]
    \centering\small
    \begin{tabular}{c|c|c|c}
    \toprule
        \multirow{2}{*}{Datasets} & {Zero-} & Search- & Gains \\
        & shot & Adaptor &(\%) \\
        \midrule
        ToolE - single tool  &  0.5292  & \textbf{0.8321} & 57.2\% \\
        ToolBench - I1 &  0.6289  & \textbf{0.7320} & 16.4\% \\
        ToolBench - I2 & 0.5054 & \textbf{0.6774} & 34.0\%\\
        ToolBench - I3 & 0.5833 & \textbf{0.7917} & 35.7\%\\
        \bottomrule
    \end{tabular}
    \caption{Tool retrieval performance with Search-Adaptor on top of Google's LLM based embedding API (gecko@latest) in terms of NDCG@1 metric.}
    \label{tab:tool_retrieval_result}
\end{table}

In this subsection, we further extend Search-Adaptor to tool retrieval application \cite{patil2023gorilla} where agents choose which actions to perform to automate execution of a task given the input query. 
The objective of tool retrieval is to retrieve the proper tools for the new query based on the descriptions of tools. 
On two datasets, ToolE \cite{huang2023metatool}, ToolBench \cite{qin2023toolllm}, we study the potential of Search-Adaptor to improve the tool retrieval performance, that would yield superior agents. 
Table~\ref{tab:tool_retrieval_result} shows that with Search-Adaptor, significant retrieval performance improvements, 15-50\%, are obtained.

\section{Discussions}\label{sec:discussion}
\subsection{Ablation studies}
Search-Adaptor proposes multiple innovations to improve the adaptation performance. 
We quantify the contributions of proposed constituents to the retrieval performance on various datasets with as ST5-Base as the base embedding model, with the results in Table~\ref{tab:ablation_study}.
We consider various modifications to Search-Adaptor: (i) altering the architecture, (ii) applying different regularizations, and (iii) applying different losses.
Using different losses yields the largest performance degradation, underlining the importance of the proposed ranking loss.
Aside from the losses, if we use separate adapters for query and corpus, it also yields a noticeable performance drop.
This shows the importance of `shared embedding space' between the query and corpus for retrieval. 
The skip connection and the two regularization functions bring additional performance gains but their impact is lower than the ranking losses.
Table~\ref{tab:ablation_study}(c) shows the impact of the proposed ranking loss in comparison to alternatives:
(i) point-wise sigmoid cross entropy, (ii) contrastive loss \citep{izacard2021unsupervised}, (iii) softmax cross entropy \citep{bruch2019analysis} and (iv) RankNet loss \citep{burges2005learning}. 
With the proposed ranking loss of the original Search-Adaptor, significant outperformance is observed, compared to the alternative ranking losses.

\subsection{Small LLMs with Search-Adaptor outperform zero-shot large LLMs}
One bottleneck for real-world LLM deployment can be the prediction latency and cost, that are highly dependent on the LLM model size. 
We demonstrate that Search-Adaptor can achieve better or comparable retrieval performance even with much smaller LLMs, compared to to zero-shot retrieval with larger LLMs. 

\begin{table}[h!]
    \centering\small
    \begin{tabular}{c|c|c||c|c}
    \toprule
        LLMs & \multicolumn{2}{|c||}{ST5-Base} & \multicolumn{2}{|c}{ST5-Large}  \\
        \midrule
        Datasets & Zero- & Search- & Zero- & Search- \\
        & shot & Adaptor & shot & Adaptor \\
        \midrule
        NFCorpus & 0.3100 & \textbf{0.3258} & 0.3354 & \textbf{0.3410}  \\
        SciFact & 0.5237 & \textbf{0.7255} & 0.5801 & \textbf{0.7530}  \\
        Arguana & 0.3646 & \textbf{0.5501} & 0.2662 & \textbf{0.4770}  \\
        SciDocs & 0.1393 & \textbf{0.1657} & 0.1618 & \textbf{0.1850}  \\
        FiQA & 0.4064 & \textbf{0.4416} & 0.4785 & \textbf{0.5028}  \\
        Trec-covid  & 0.5990 & \textbf{0.6986} & 0.6471 & \textbf{0.7082} \\
        Touche  & 0.2291 & \textbf{0.3393} & 0.2624 & \textbf{0.3408} \\
        Quora  & 0.7484 & \textbf{0.8664} & 0.7560 & \textbf{0.9705}  \\
        \midrule
        Average & 0.4151 & \textbf{0.5141} & 0.4607 & \textbf{0.5223} \\
        \bottomrule
    \end{tabular}
    \caption{The performance of Search-Adaptors when applied on Sentence-T5 models, (i) ST5-Base (110M parameters) and (ii) ST5-Large (335M parameters), in terms of nDCG@10 metric.}
    \label{tab:t5_result}
\end{table}


As shown in Table~\ref{tab:t5_result}, Search-Adaptor with ST5-Base model (110M parameters) performs much better than ST5-Large (335M parameters). 
Search-Adaptor can achieve better results with much smaller encoders, positioning it as an effective distillation mechanism to significantly decrease the serving cost and latency of retrieval systems.
It also reiterates the benefits of Search-Adaptor being model agnostic.

\section{Conclusions}\label{sect:conclusion}
In this paper, we focus on pushing the capabilities of LLMs for information retrieval and search. 
We propose a canonical efficient adaptation method, Search-Adaptor, that can also be applied to LLMs even with inference-only access. 
Search-Adaptor is a low-cost tuning method that brings significant and consistent improvements in retrieval performance across diverse regimes of training data size, encoder type, and corpus set. 
This is enabled by the judicious design of its adaptor module, along with training objectives and approaches.
We have also studied the extension of Search-Adaptor to multimodal learning and tool use scenarios, highlighting the importance of embedding customization for such applications. 

\section{Limitations and Future Works}
Important future directions might include generalizing the propose adaptation method to include partial tuning of the embedding models as a way to increase trainable degrees of freedom; extensions to embedding tasks beyond retrieval; and extensions to multimodal learning with many modalities. 
There is no specific risk of the proposed method other than the general risks of tuning methods that they can lead to overfitting to certain tasks and they can absorb the biases present in the target tuning data. 


\bibliography{acl_latex}

\onecolumn
\appendix

\section{Data Statistics}\label{appx:data}

\subsection{BEIR datasets}

\begin{table}[h!]
    \small
    \centering
    \begin{tabular}{c|c|c|c}
    \toprule
        Datasets & The number of train pairs  & The number of test pairs& The number of corpus \\
        \midrule
        NFCorpus & 110575 & 12334 & 3.6K \\
        SciFact & 919 & 339 & 5K \\
        Arguana & 703 & 703 & 8.67K \\
        SciDocs & 14972 & 14956 & 25K \\
        FiQA & 14166 & 1706 & 57K \\
        Trec-Covid & 35460 & 30876 & 171K \\
        Touche & 1077 & 1137 & 382K \\
        Quora & 7626 & 15675 & 523K \\
        NQ & 2097 & 2104 & 2.68M \\
        DBPedia & 5673 & 43515 & 4.63M \\
        HotPotQA & 170000 & 14810 & 5.23M \\
        Fever & 140085 & 7937 & 5.42M \\
        Climate-fever & 2299 & 2382 & 5.42M\\
        MSMarco & 532751 & 9260 & 8.84M \\
        \bottomrule
    \end{tabular}
    \caption{The statistics of the BEIR datasets (sorted by the number of corpus).}
    \label{tab:data_stats}
\end{table}

\subsection{MIRACL datasets}

\begin{table}[h!]
    \small
    \centering
    \begin{tabular}{c|c|c|c}
    \toprule
        Datasets & The number of train pairs  & The number of test pairs& The number of corpus \\
        \midrule
        Yoruba (yo) & 959 & 229 & 49043 \\
        Swahilli (sw) & 9359 & 5092 & 131924 \\
        Bengali (bn) & 16754 & 4206 & 297265 \\
        Hindi (hi) & 11668 & 3494 & 506264 \\
        Telugu (te) & 18608 & 1606 & 518079 \\
        Thai (th) & 21293 & 7573 & 542166 \\
        \midrule
        Indonesian (id) & 41358 & 9668 & 1446315 \\
        Korean (ko) & 12767 & 3057 & 1486752 \\
        Finnish (fi) & 20350 & 12008 & 1883509 \\
        Arabic (ar) & 25382 & 29197 & 2061414 \\
        Persian (fa) & 21844 & 6571 & 2207172 \\
        Chinese (zh) & 13113 & 3928 & 4934368 \\
        \midrule
        Japanese (ja) & 34387 & 8354 & 6953614\\
        Russian (ru) & 33921 & 13100 & 9543918 \\
        Spanish (es) & 21531 & 6443 & 10373953 \\
        French (fr) & 11426 & 3429 & 14636953 \\
        Germany (de) & 2526 & 628 & 15866222 \\
        \bottomrule
    \end{tabular}
    \caption{The statistics of the MIRACL datasets (sorted by the number of corpus).}
    \label{tab:miracl_data_stats}
\end{table}

\subsection{Fashion-200K datasets}

\begin{table}[h!]
    \small
    \centering
    \begin{tabular}{c|c|c|c}
    \toprule
        Datasets & The number of train pairs  & The number of test pairs& The number of corpus \\
        \midrule
        Dresses & 15127 & 1567 & 72376 \\
        Jackets & 8105 & 1511 & 71118 \\
        Pants & 9264 & 1758 & 74470 \\
        Skirts & 6822 & 1247 & 47931 \\
        Tops & 13809 & 2536 & 72444 \\
        \bottomrule
    \end{tabular}
    \caption{The statistics of the Fashion-200K datasets.}
    \label{tab:fashion_mnist_data_stats}
\end{table}

\newpage
\subsection{Tool retrieval datasets}

\begin{table}[h!]
    \small
    \centering
    \begin{tabular}{c|c|c|c}
    \toprule
        Datasets & The number of train pairs  & The number of test pairs& The number of corpus \\
        \midrule
        ToolE - single tool & 16440 & 4110 & 199 \\
        ToolBench - I1 & 87322 & 97 & 10439 \\
        ToolBench - I2 & 84722 & 93 & 13142 \\
        ToolBench - I3 & 25155 & 96 & 1605 \\
        \bottomrule
    \end{tabular}
    \caption{The statistics of the tool retrieval datasets (ToolE and ToolBench).}
    \label{tab:tool_retrieval_data_stats}
\end{table}

\section{Metrics}\label{appx:metric}
For tasks that involve retrieving information, normalized discounted cumulative gain (nDCG) \citep{jarvelin2002cumulated} is a standard metric for evaluating performance. To define nDCG, we first consider discounted cumulative gain (DCG):
\begin{equation*}
    DCG(y, s) = \sum_{i}\frac{2^{y_i}}{\log_2(\text{rank}(s_i)+1)},
\end{equation*}
where $s$ is the relevance score computed by the model and $y$ is the ground truth label.
nDCG is then defined as $nDCG(y, s) = \frac{DCG(y, s)}{DCG(y, y)}$, where the denominator assumes the optimal case where the ranking of the scores ($s$) are exactly the same as the ranking of the ground truth label ($y$). nDCG@k is a widely used variation of nDCG where only the top $k$ scores are considered. In this paper, we use nDCG@10 as the main retrieval metric.

\section{Hyper-parameters}\label{appx:hyperparams}
We summarize the hyper-parameters used to train Search-Adaptor. In all experiments, we utilize the fixed hyper-parameters (except $\alpha, \beta$) that enable applying Search-Adaptor without extensive hyper-parameter tuning. We use multi-layer perceptron as the adaptor architecture for both the encoder and the predictor.

\begin{table}[h!]
    \small
    \centering
    \begin{tabular}{l|l}
    \toprule
        Hyper-parameters & Fixed values \\
        \midrule
        Recovery loss coefficient ($\alpha$) & $\{0.0,0.1,1.0\}$\\
        Prediction loss coefficient ($\beta$) & $\{0.0,0.01,0.1\}$ \\
        Batch size for training & 128 \\
        Maximum number of training iterations & 2000 \\
        Patience for early stopping & 125 \\
        Learning rates & 0.001 \\
        Optimizer & Adam \\
        Negative pair subsampling ratio (compared with positive pairs) & 10 \\
    \bottomrule
    \end{tabular}
    \caption{Hyper-parameters used to train Search-Adaptor in all experiments.}
    \label{tab:hyperparams}
\end{table}

\newpage
\section{Additional Experiments}\label{appendix:additional_base_embedding_model}
We include the additional results of Search-Adaptor with GTR-Large\footnote{\url{https://huggingface.co/sentence-transformers/gtr-t5-large}} \cite{ni2021large} and Condenser-Retriever \footnote{\url{https://huggingface.co/Luyu/co-condenser-marco-retriever}} \cite{gao2021condenser} as the base embedding models. As can be seen in Table.~\ref{tab:gtr}, the results are consistent with the above results that Search-Adaptor shows consistent and significant improvements on top of both GTR-Large and Condenser-Retriever models.
For the Condenser-Retriever model, we apply pooling and normalization on the token embeddings to extract the final text embeddings.

\begin{table}[h!]
    \centering
    \small
    \begin{tabular}{c|c|c|c||c|c|c}
    \toprule
        \multirow{3}{*}{Datasets} & \multicolumn{3}{c||}{GTR-Large Model}  & \multicolumn{3}{|c}{Condenser-Retriever Model} \\
        \cmidrule{2-7}
        & Zero-shot & Search-Adaptor & Gains (\%) & Zero-shot & Search-Adaptor & Gains (\%)  \\
        \midrule
        NFCorpus & 0.3148 & \textbf{0.3242}  & 2.99\% & 0.0882 & \textbf{0.2506} & 184.13\%\\
        SciFact & 0.5331 & \textbf{0.7469}  & 40.11\% &0.2182 & \textbf{0.6783} & 210.86\% \\
        Arguana & 0.5139 & \textbf{0.6360}  & 23.76\% & 0.2744 & \textbf{0.3757} & 36.92\% \\
        SciDocs & 0.1657 & \textbf{0.1687}  & 1.81\% & 0.0659 & \textbf{0.1215} & 84.37\% \\
        FiQA & 0.4069 & \textbf{0.4265}  & 4.82\% & 0.0775 & \textbf{0.2445} & 215.48\% \\
        Trec-Covid & 0.6912 & \textbf{0.7481}  & 8.23\% & 0.3416 & \textbf{0.5769} & 68.88\% \\
        Touche & 0.2723 & \textbf{0.3227} & 18.51\% & 0.0623 & \textbf{0.1928} & 8.34\%  \\
        Quora & 0.8428 & \textbf{0.8795} & 4.35\%  & 0.7937 & \textbf{0.8599} & 8.34\% \\
        \midrule
        Average & 0.4676 & \textbf{0.5315} & 13.68\% & 0.2402 & \textbf{0.4125} & 71.72\%\\
        \bottomrule
    \end{tabular}
    \caption{Performance improvements with Search-Adaptor on top of GTR-Large and Condenser-Retriever embedding models.}
    \label{tab:gtr}
\end{table}

\newpage
\section{Qualitative Analysis}\label{appendix:qualitative_analysis_sa}
To understand the impact of Search-Adaptor, we first analyze the cosine similarity between query and corpus, before and after Search-Adaptor training. 

\begin{figure*}[h!]
\centering
\begin{subfigure}{0.49\textwidth}
  \includegraphics[width=\linewidth]{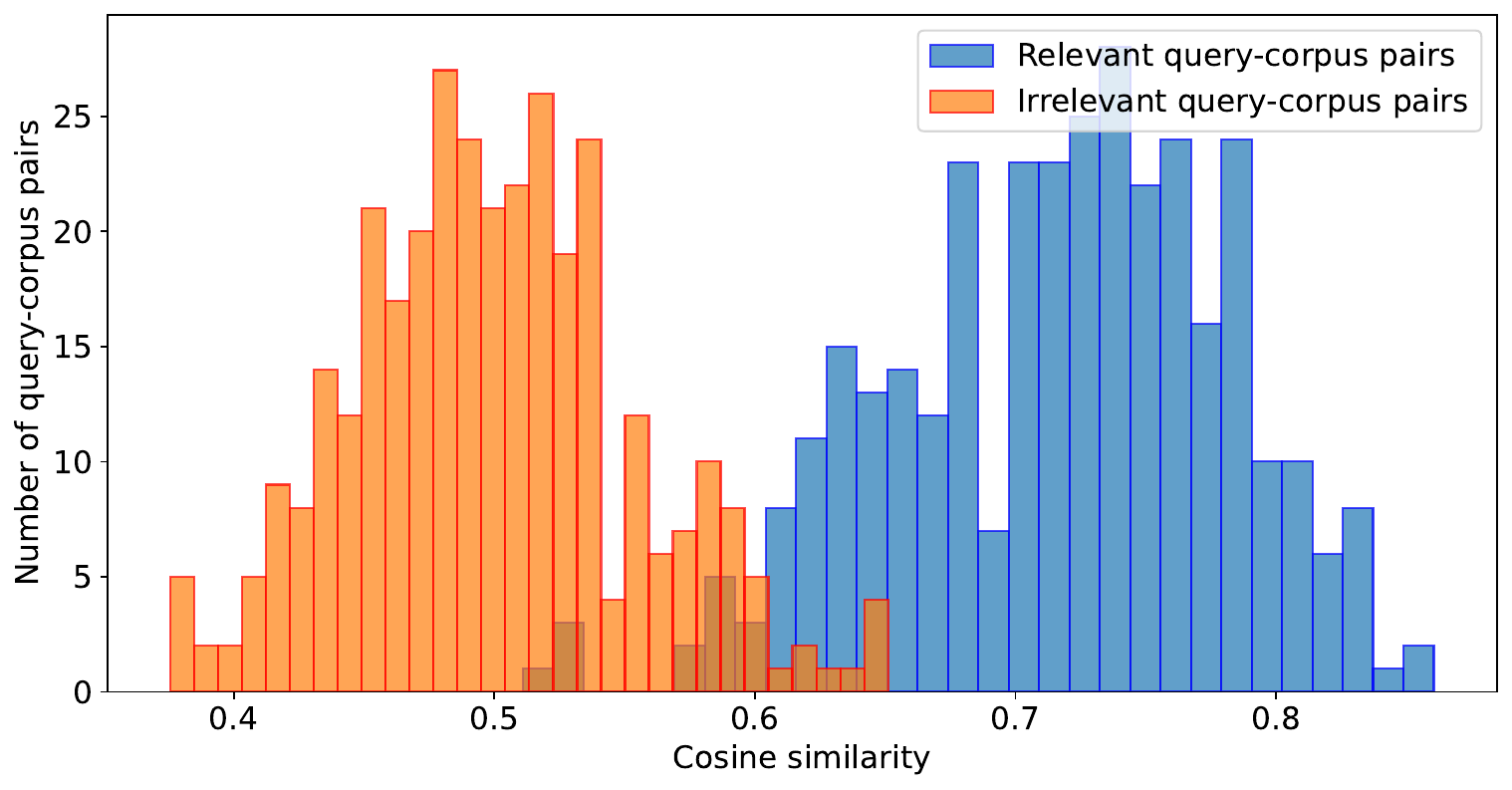}
  \caption{Score distribution before Search-Adaptor}
\end{subfigure}
\begin{subfigure}{0.49\textwidth}
  \includegraphics[width=\linewidth]{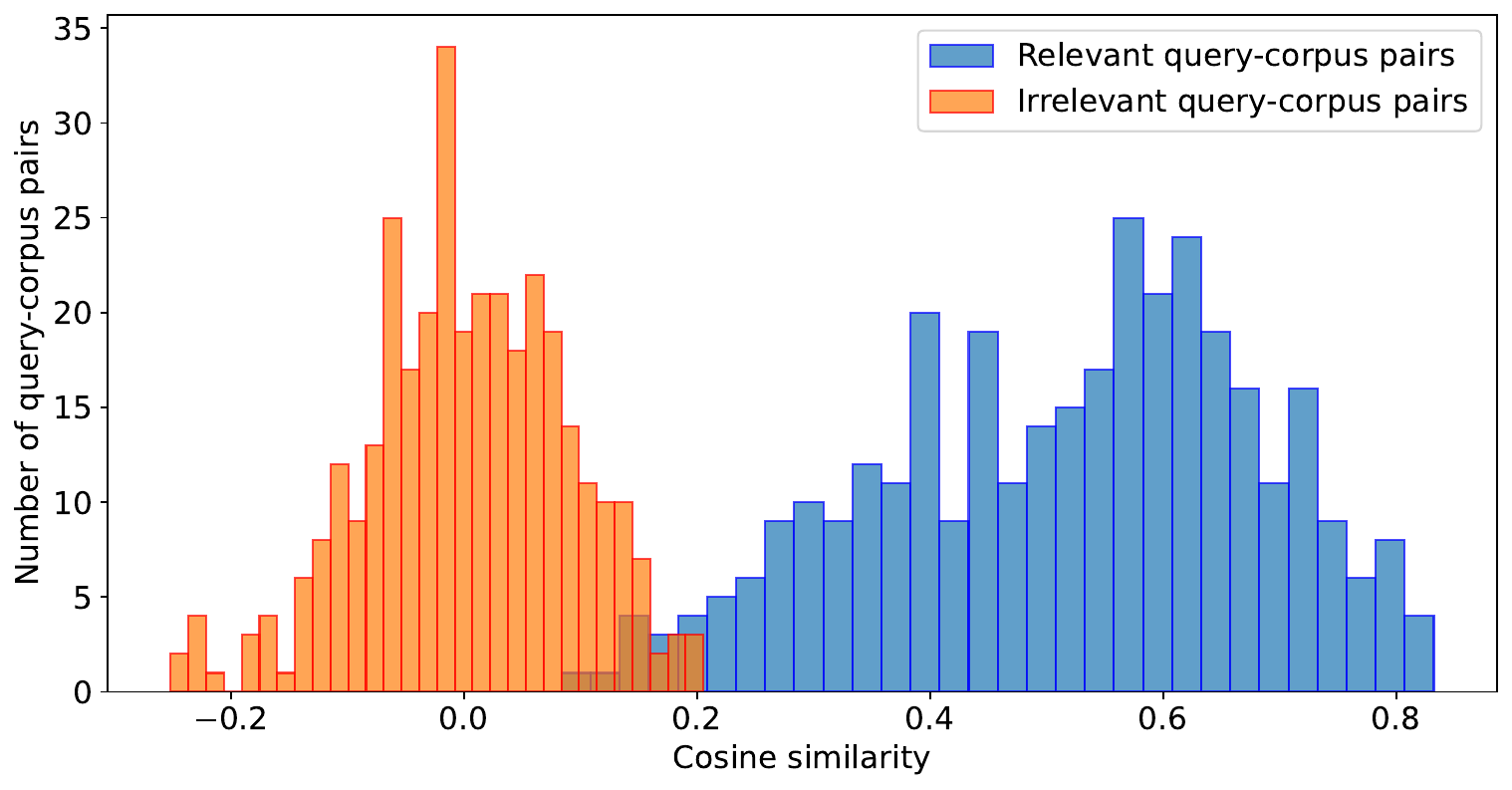}
  \caption{Score distribution after Search-Adaptor}
\end{subfigure}
\caption{Cosine similarity score distributions before and after Search-Adaptor.}
\label{fig:analysis1}
\end{figure*}

As can be seen in Fig.~\ref{fig:analysis1}, after Search-Adaptor training, the distribution differences between relevant and irrelevant pairs’ cosine similarity are larger which means that we can identify the relevant corpus per each query better.

To further understand the distribution difference of query and corpus embeddings before and after Search-Adaptor training, we plot tSNE graphs of them.

\begin{figure*}[h!]
\centering
\begin{subfigure}{0.49\textwidth}
  \includegraphics[width=\linewidth]{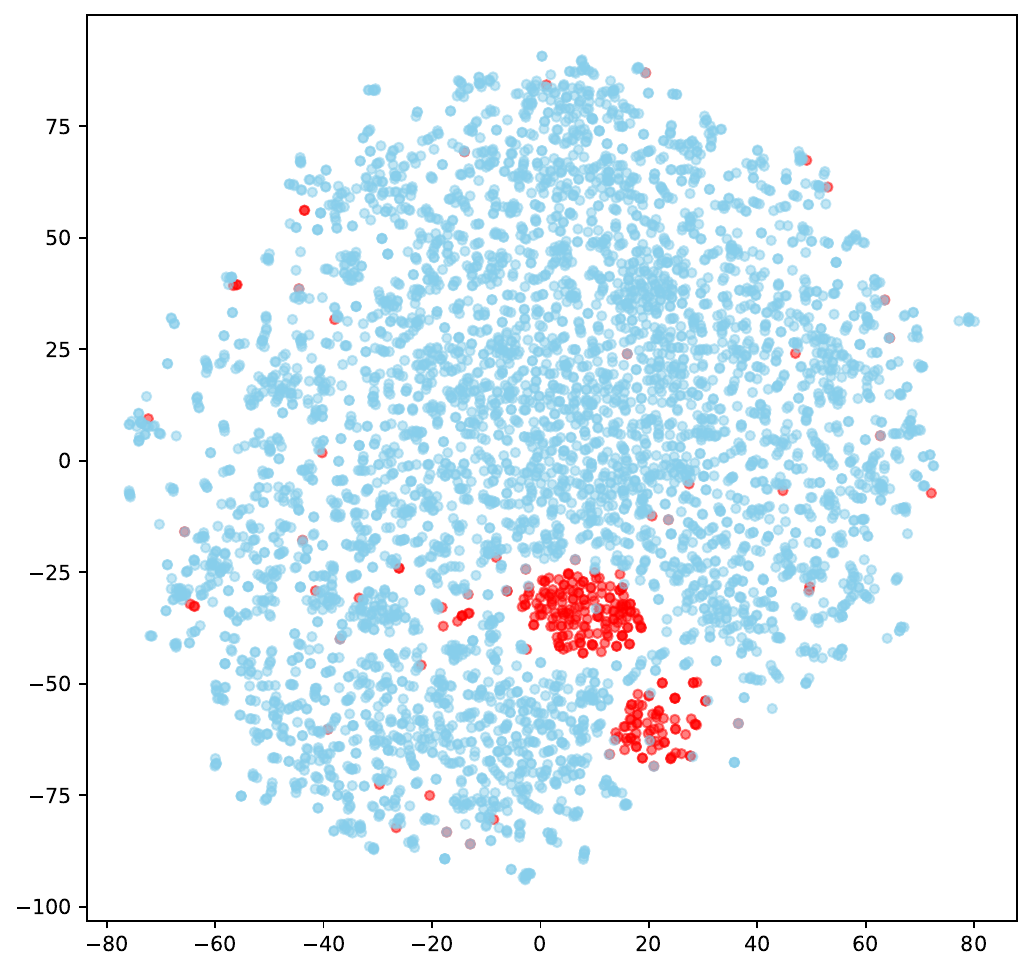}
  \caption{tSNE analysis before Search-Adaptor}
\end{subfigure}
\begin{subfigure}{0.49\textwidth}
  \includegraphics[width=\linewidth]{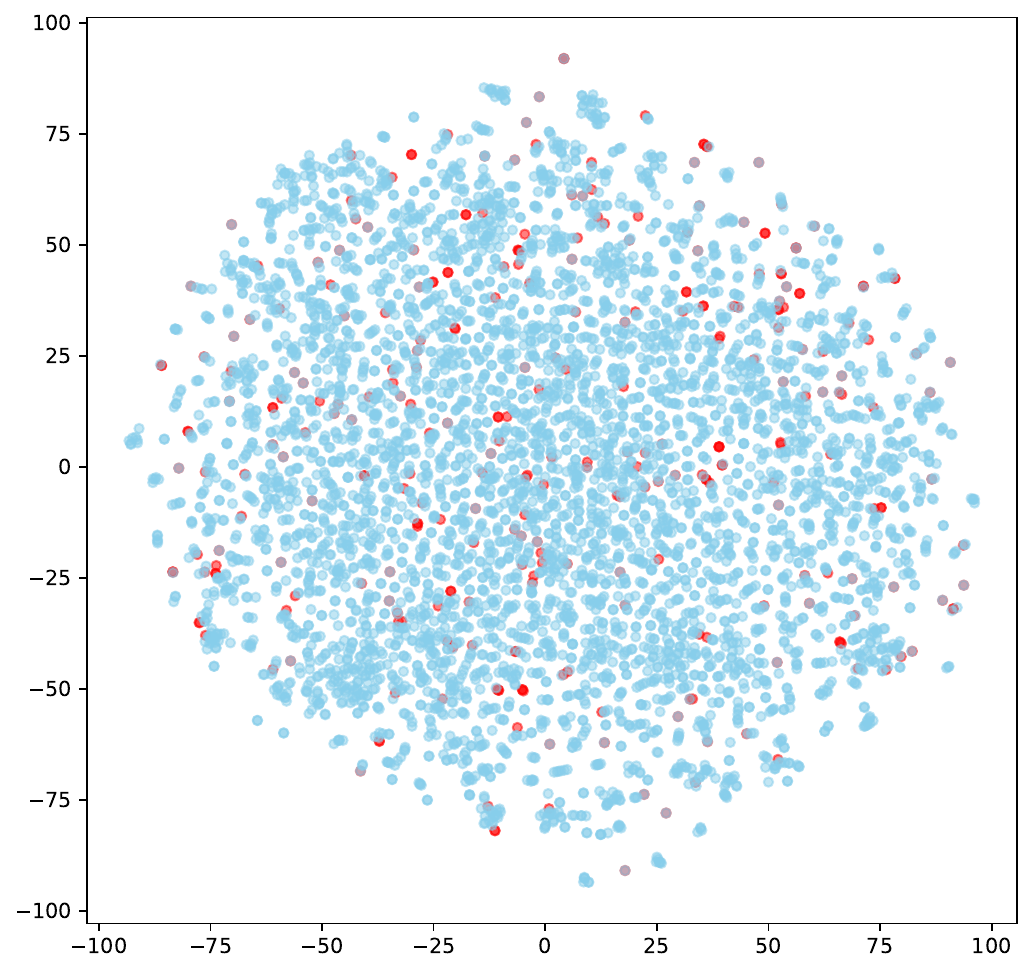}
  \caption{tSNE analysis after Search-Adaptor}
\end{subfigure}
\caption{tSNE distributions before and after Search-Adaptor. Red represents query embedding and blue represents corpus embedding.}
\label{fig:analysis2}
\end{figure*}

Fig.~\ref{fig:analysis2} shows the impact of Search-Adaptor. The left figure shows that the original query and corpus embeddings are quite distinct. Most query embeddings are located in the restricted region. On the other hand, after training with Search-Adaptor, query embedding distribution is observed to better overlap with the corpus embedding distribution, which could result in more robust retrieval. 

\newpage
We further investigate the success and failure cases of Search-Adaptor in comparison to the zero-shot baseline. Bold represents the relevant corpus to the query.
\begin{table*}[h!]
    \small
    \centering
    \resizebox{0.99\textwidth}{!}{
    \begin{tabular}{p{0.2\textwidth}||p{0.4\textwidth}||p{0.4\textwidth}}
    \toprule
        \textbf{Query} & \textbf{Baseline Retrieval} & \textbf{Search-Adaptor Retrieval} \\
    \midrule
        {Suboptimal nutrition is not predictive of chronic disease} & Maternal and child undernutrition: consequences for adult health and human capital & \textbf{Global, regional, and national comparative risk assessment of 79 behavioural, environmental and occupational, and metabolic risks or clusters of risks, 1990–2015: a systematic analysis for the Global Burden of Disease Study 2015}\\
        \cmidrule{2-3}
        & Effect of women's nutrition before and during early pregnancy on maternal and infant outcomes: a systematic review. & Dietary quality among men and women in 187 countries in 1990 and 2010: a systematic assessment \\
        \cmidrule{2-3}
        & Dietary quality among men and women in 187 countries in 1990 and 2010: a systematic assessment & Biomarkers of endothelial dysfunction and risk of type 2 diabetes mellitus. \\
        \midrule
        The PRR MDA5 is a sensor of RNA virus infection. & Ribose 2-O-methylation provides a molecular signature for the distinction of self and non-self mRNA dependent on the RNA sensor Mda5 & \textbf{Immune signaling by RIG-I-like receptors.} \\
        \cmidrule{2-3}
        & \textbf{Immune signaling by RIG-I-like receptors.} & Ribose 2-O-methylation provides a molecular signature for the distinction of self and non-self mRNA dependent on the RNA sensor Mda5 \\
        \cmidrule{2-3}
        & RIG-I-mediated antiviral responses to single-stranded RNA bearing 5'-phosphates. & RIG-I-mediated antiviral responses to single-stranded RNA bearing 5'-phosphates.\\
        \midrule
        A deficiency of vitamin B12 increases blood levels of homocysteine. & Preventing coronary heart disease: B vitamins and homocysteine. & \textbf{Folic acid improves endothelial function in coronary artery disease via mechanisms largely independent of homocysteine lowering.}\\
        \cmidrule{2-3}
        & Effect of homocysteine lowering on mortality and vascular disease in advanced chronic kidney disease and end-stage renal disease: a randomized controlled trial. & \textbf{Randomized trial of folic acid supplementation and serum homocysteine levels.} \\
        \cmidrule{2-3}
        & Hyperhomocysteinemia and atherosclerotic vascular disease: pathophysiology, screening, and treatment. off. & The effect of folic acid supplementation on plasma homocysteine in an elderly population.\\
    \bottomrule 
    \end{tabular}
    }
    \caption{Success cases: Examples of query and top-3 retrieved documents where relevant documents are ranked higher in Search-Adaptor in comparison to baseline. Top-3 retrieved documents' titles are listed above.}
    \label{tab:success_cases}
\end{table*}

\newpage
\begin{table*}[t!]
    \centering
    \small
    \resizebox{0.99\textwidth}{!}{
    \begin{tabular}{p{0.3\textwidth}||p{0.35\textwidth}||p{0.35\textwidth}}
    \toprule
        \textbf{Query} & \textbf{Baseline Retrieval} & \textbf{Search-Adaptor Retrieval} \\
    \midrule
        {Antibiotic induced alterations in the gut microbiome reduce resistance against Clostridium difficile} & \textbf{Antibiotic-induced shifts in the mouse gut microbiome and metabolome increase susceptibility to Clostridium difficile infection} & Precision microbiome reconstitution restores bile acid mediated resistance to Clostridium difficile \\
        \cmidrule{2-3}
        & Precision microbiome reconstitution restores bile acid mediated resistance to Clostridium difficile& \textbf{Antibiotic-induced shifts in the mouse gut microbiome and metabolome increase susceptibility to Clostridium difficile infection}\\
        \cmidrule{2-3}
        & Role of gut commensal microflora in the development of experimental autoimmune encephalomyelitis.& Microbiome-driven allergic lung inflammation is ameliorated by short-chain fatty acids\\
        \midrule
        The genomic aberrations found in matasteses are very similar to those found in the primary tumor.&\textbf{Evolution of metastasis revealed by mutational landscapes of chemically induced skin cancers} &Intratumor heterogeneity and branched evolution revealed by multiregion sequencing. \\
        \cmidrule{2-3}
        & Molecular characterization of endometrial cancer: a correlative study assessing microsatellite instability, MLH1 hypermethylation, DNA mismatch repair protein expression, and PTEN, PIK3CA, KRAS, and BRAF mutation analysis. &Diverse tumorigenic pathways in ovarian serous carcinoma. \\
        \cmidrule{2-3}
        &Deregulated DNA polymerase beta induces chromosome instability and tumorigenesis. &\textbf{Evolution of metastasis revealed by mutational landscapes of chemically induced skin cancers} \\
        \midrule
        Incidence rates of cervical cancer have increased due to nationwide screening programs based primarily on cytology to detect uterine cervical cancer.&\textbf{Mass screening programmes and trends in cervical cancer in Finland and the Netherlands.} &The effect of mass screening on incidence and mortality of squamous and adenocarcinoma of cervix uteri. \\
        \cmidrule{2-3}
        &The effect of mass screening on incidence and mortality of squamous and adenocarcinoma of cervix uteri. &\textbf{Mass screening programmes and trends in cervical cancer in Finland and the Netherlands.} \\
        \cmidrule{2-3}
        & Efficacy of human papillomavirus testing for the detection of invasive cervical cancers and cervical intraepithelial neoplasia: a randomised controlled trial.& Efficacy of human papillomavirus testing for the detection of invasive cervical cancers and cervical intraepithelial neoplasia: a randomised controlled trial.\\
    \bottomrule 
    \end{tabular}
    }
    \caption{Failure cases: Examples of query and top-3 retrieved documents where relevant documents are ranked higher in baseline in comparison to Search-Adaptor. Top-3 retrieved documents' titles are listed above.}
    \label{tab:failure_cases}
\end{table*}

As can be seen in Table.~\ref{tab:success_cases} and \ref{tab:failure_cases}, in failure cases, Search-Adaptor still can retrieve the relevant corpus in the top-3 corpus but the ranking is lower than the baseline. For the success cases, Search-Adaptor can retrieve the correct corpus even though the baseline is completely failed. Quantitatively, with 300 test samples, there are 9 cases where Search-Adaptor can retrieve the correct corpus in top-3 but Baseline cannot retrieve any correct corpus in top-3. But there is no case for the opposite.

\newpage
\section{Qualitative Analysis on Multimodal Data}\label{sect:qualitative_multimodal}
\begin{figure}[h!]
    \centering
    \includegraphics[width=0.94\textwidth]{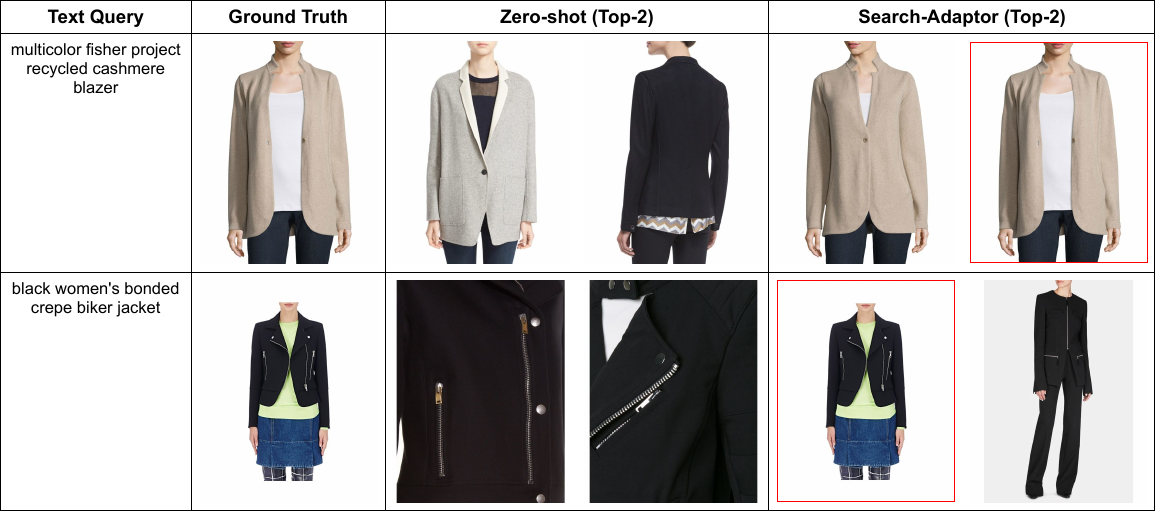}\\
    \includegraphics[width=0.94\textwidth]{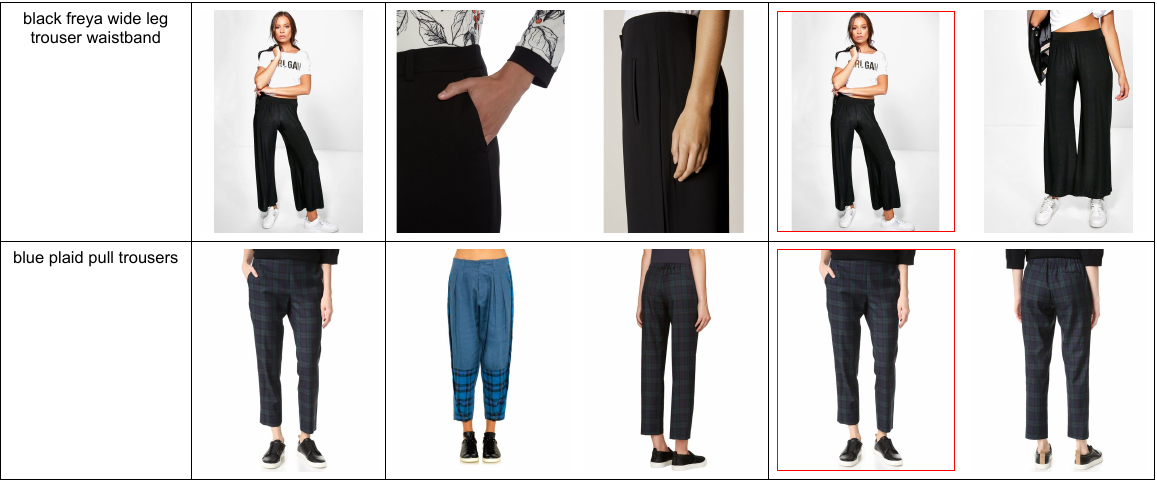}\\
    \includegraphics[width=0.94\textwidth]{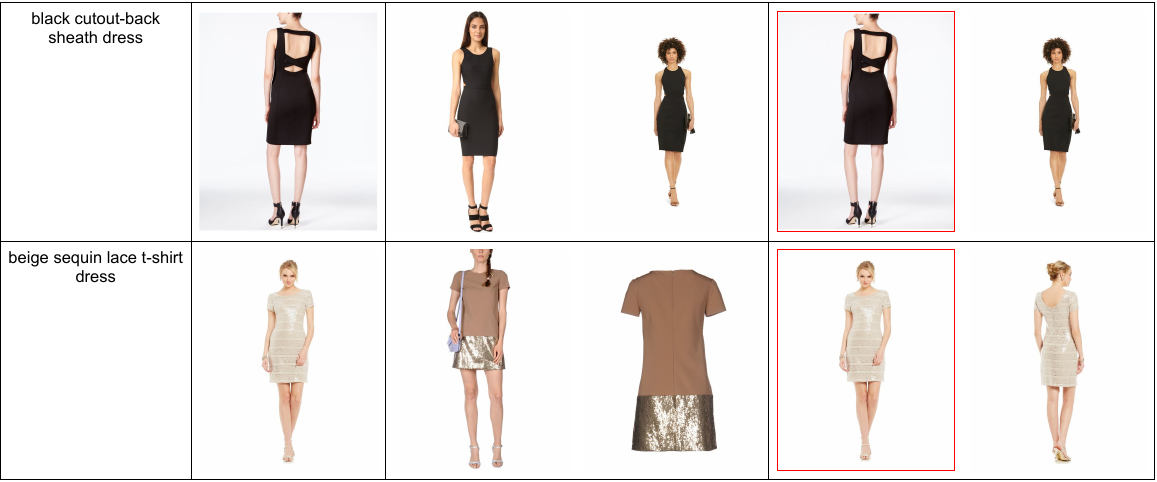}\\
    \caption{Qualitative analyses of text to image retrieval with Search-Adaptor using Fashion-200K data.}
    \label{fig:qualitative_result_multimodal}
\end{figure}
Fig.~\ref{fig:qualitative_result_multimodal} shows 6 examples in Fashion-200K datasets where Search-Adaptor makes better retrieved output than the zero-shot baseline (Google Cloud's multimodal embedding API). First column represents the given text query. Second column represents the ground truth relevant image for the given text query. Third column shows the top-2 retrieved outputs based on the zero-shot baseline. Last column shows the top-2 retrieved outputs based on Search-Adaptor.
Note that the ground truths are included in top-2 retrieved outputs by Search-Adaptor but not included in the baseline.

\end{document}